\documentclass[conference]{IEEEtran}
\IEEEoverridecommandlockouts
\usepackage{cite}
\usepackage{amsmath,amssymb,amsfonts}
\usepackage{algorithmic}
\usepackage{graphicx}
\usepackage{textcomp}
\usepackage{xcolor}
\usepackage{hyperref}
\usepackage{tabularx}
\def\BibTeX{{\rm B\kern-.05em{\sc i\kern-.025em b}\kern-.08em
    T\kern-.1667em\lower.7ex\hbox{E}\kern-.125emX}}
\begin{document}

\title{A workflow for generating synthetic LiDAR datasets in simulation environments.\\

}

\author{\IEEEauthorblockN{Abhishek Phadke}
\IEEEauthorblockA{\textit{School of Engineering \& Computing} \\
\text{Christopher Newport University}\\
\text{abhishek.phadke@cnu.edu}}
\and
\IEEEauthorblockN{Shakib Mahmud Dipto}
\IEEEauthorblockA{\textit{Department of Computer Science} \\
\text{Old Dominion University}\\
\text{diptomahmud2@gmail.com}}
\and
\IEEEauthorblockN{Pratip Rana}
\IEEEauthorblockA{\textit{Department of Computer Science} \\
\text{Old Dominion University}\\
\text{prana@odu.edu}}
}

\maketitle

\begin{abstract}
This paper presents a simulation workflow for generating synthetic LiDAR datasets to support autonomous vehicle perception, robotics research, and sensor security analysis. Leveraging the CoppeliaSim simulation environment and its Python API, we integrate time-of-flight LiDAR, image sensors, and two-dimensional scanners onto a simulated vehicle platform operating within an urban scenario. The workflow automates data capture, storage, and annotation across multiple formats (PCD, PLY, CSV), producing synchronized multimodal datasets with ground truth pose information. We validate the pipeline by generating large-scale point clouds and corresponding RGB and depth imagery. The study examines potential security vulnerabilities in LiDAR data, such as adversarial point injection and spoofing attacks, and demonstrates how synthetic datasets can facilitate the evaluation of defense strategies. Finally, limitations related to environmental realism, sensor noise modeling, and computational scalability are discussed, and future research directions, such as incorporating weather effects, real-world terrain models, and advanced scanner configurations, are proposed. The workflow provides a versatile, reproducible framework for generating high-fidelity synthetic LiDAR datasets to advance perception research and strengthen sensor security in autonomous systems. Documentation and examples accompany this framework; samples of animated cloud returns and image sensor data can be found at this \href{https://github.com/abhishekphadke/synthetic-LiDARdata-coppeliaSim.git}{Link}
\end{abstract}


\section{Introduction}\label{1}
Light Detection and Ranging (LiDAR) is a remote sensing method that utilizes laser light to measure distances. In a typical time-of-flight (ToF) LiDAR system, short laser pulses are emitted toward a target, and a detector records the time it takes for the reflected light to return. The distance to the target, $d$, is calculated based on this round-trip time interval, given the known speed of light. The LiDAR unit calculates the range using the formula $d = c\times \Delta t /2$, where $c$ represents the speed of light and $\Delta t$ denotes the round-trip time. Alternatives such as frequency-modulated continuous wave (FMCW) LiDAR also exist, emitting a continuous chirped laser and measuring the return signal's frequency shift. While FMCW LiDAR can simultaneously determine distance and relative velocity by mixing the returned light with the outgoing signal, due to its simpler design and lower cost, ToF has become the dominant approach in most current LiDAR systems \cite{b1}. Technological advancements have led to other lidar mechanisms, such as flash LiDAR. For all types of LiDAR, sensitive photodetectors and timing electronics generally capture the returning light, which together determine distances with high precision, often within a few centimeters or better in real-time \cite{b2}.
\\
LiDAR produces precise three-dimensional measurements; hence, it has broad applications in environmental and geospatial fields. In forestry and ecology, airborne LiDAR is used to map forest structures, measure tree heights, and estimate biomass \cite{b3, b4}. Single-wavelength LiDAR surveys can capture horizontal biomass distribution and vertical canopy profiles in research plots \cite{b5}. The ability of laser pulses to penetrate between leaves and record multiple return echoes enables detailed models of vegetation layers and underlying ground terrain, which is valuable for forest management and wildlife habitat assessment. In general, environmental monitoring and earth science, LiDAR on aircraft or drones generates high-resolution digital elevation models (DEMs) for terrain analysis, hydrological modeling, and land-use planning. Repeated LiDAR scans can detect landscape changes like erosion or landslides, supporting hazard monitoring. LiDAR has also revolutionized archaeology and cultural heritage surveying \cite{b6,b7}. High-density laser scans from airplanes can unveil archaeological features hidden by dense vegetation or soil. In Central and South America, airborne LiDAR mapping has revealed extensive pre-Hispanic settlement patterns beneath rainforest canopies that were not visible to surveys on foot \cite{b6}. By producing a bare-earth model, LiDAR exposes subtle earthworks, ancient building foundations, and roads with great accuracy, thus aiding discoveries and site preservation. Another domain is urban planning and infrastructure. LiDAR is employed to create detailed 3D models of cityscapes, buildings, and transportation networks, which assist in city planning, construction, and smart-city development \cite{b9, b10}. LiDAR has been widely adopted in urban mapping due to its high spatial accuracy. It can capture exact structure heights and layouts, useful for zoning analyses, line-of-sight studies, and modeling urban flood scenarios. Municipalities use mobile and aerial LiDAR to survey roads and bridges, identifying structural deformations or encroaching vegetation. In addition, LiDAR contributes to precision agriculture and geology. In summary, the rich 3D point clouds from LiDAR enable applications ranging from generating city and terrain models to monitoring forests and uncovering archaeological sites, making it a versatile tool across environmental sciences, engineering, and heritage conservation \cite{b5}.
\\
LiDAR technology has seen rapid adoption in fields like autonomous driving and environmental mapping due to its precision in generating 3D spatial data. However, it suffers from key limitations: environmental factors like fog and rain degrade performance, while occlusion and surface reflectivity affect data quality. Moreover, real-world LiDAR datasets are expensive to collect, limited in geographic scope, and often withheld due to strategic sensitivity. Synthetic datasets address accessibility but often lack realism due to missing noise models, simplified textures, and the absence of weather effects. Security remains another pressing concern. LiDAR systems are vulnerable to spoofing and adversarial point injection, with deep learning-based perception models easily deceived. This work is another step towards addressing these issues by demonstrating a high-fidelity LiDAR data generation workflow using CoppeliaSim. It supports multimodal data (3D/2D LiDAR, RGB, depth), synchronized capture, pose-annotated outputs, and planned support for security evaluation and noise modeling.

\section{Background and Related Work}\label{2}
This section is arranged in the following manner. The study first outlines applications of LiDAR data for vehicles and discusses some common general and security issues related to LiDAR sensors and LiDAR data. Current relevant literature that addresses these issues is referenced.

\subsection{Applications of LiDAR data for vehicles}\label{2a}
\begin{figure}[h]
    \centering
    \includegraphics[width=\linewidth]{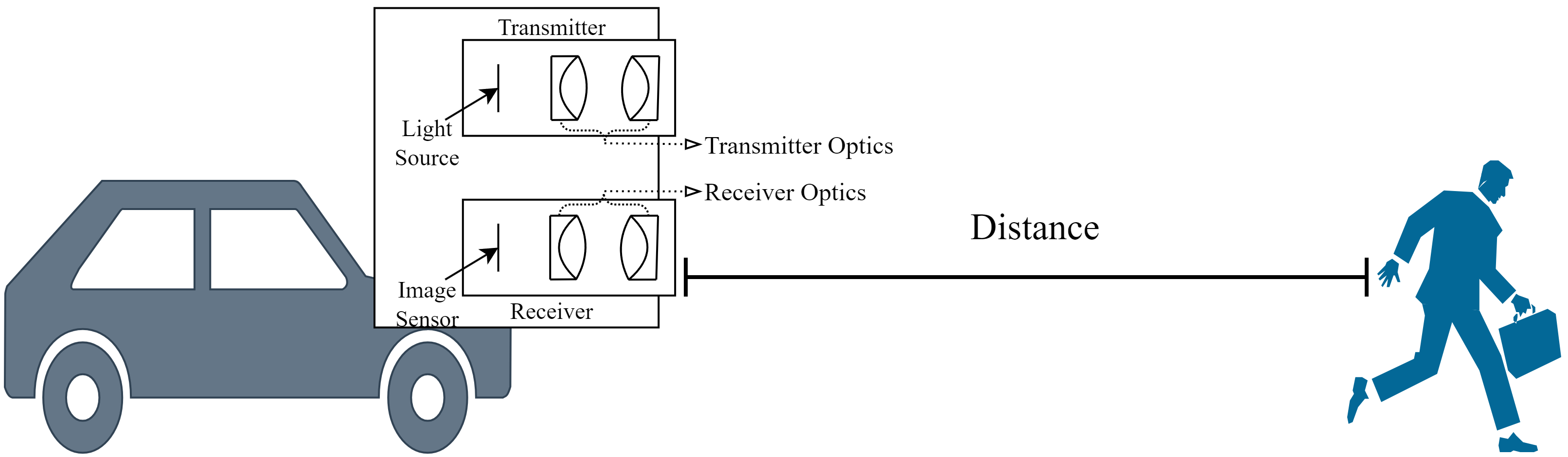}
    \caption{LiDAR and image sensor placement on autonomous driving capable vehicles.}
    \label{fig1}
\end{figure}

LiDAR has emerged as a key sensor in modern vehicles for autonomous driving \cite{b11} and advanced driver-assistance systems (ADAS). It provides a high-resolution depth perception of a car’s surroundings, which is critical for detecting obstacles, pedestrians, and other vehicles in real-time. A LiDAR unit rapidly scans the environment and produces a dense point cloud, a 3D map of discrete points reflecting off surrounding objects. This real-time 3D vision gives self-driving cars an accurate understanding of object positions and shapes around them, even in low-light conditions where cameras may struggle. In the 2020s, many high-end cars began integrating LiDAR to enhance their perception capabilities, as LiDAR sensors can deliver precise ranging information that complements cameras and radar \cite{b1}. A LiDAR-based system can reliably measure a car's distance and relative speed ahead, enabling features like adaptive cruise control and automatic emergency braking to function with improved safety margins. In an autonomous vehicle’s sensor suite, LiDAR typically works alongside cameras (for color and texture) and radar (for long-range and velocity data) in a fusion approach. A typical automotive LiDAR system includes a laser emitter/scanner and photoreceiver, plus processing circuits, all mounted on the vehicle. As it scans, it feeds range data to the vehicle’s perception algorithms, which identify and track objects. High-end setups may use multiple LiDAR units, such as a forward-facing primary LiDAR for long-distance view and additional side or rear LiDARs to cover blind spots, achieving a 360° field of view around the car. This rich depth data substantially improves object detection and mapping, such as distinguishing pedestrians from background objects by their 3D profile or precisely localizing the roadway and free space for path planning. \autoref{fig1} shows the genralized working of LiDAR sensor on a vehicle. 

Research has shown that LiDAR’s direct velocity measurements or frame-to-frame tracking of points can even help predict the intentions of other actors, such as detecting a cyclist’s speed and trajectory \cite{b11}. LiDAR strengthens situational awareness for autonomous driving by providing reliable distance and geometry information. Vehicles with LiDAR have demonstrated more robust performance in lane merging and obstacle avoidance tasks since the sensor can pinpoint objects within centimeters. However, current LiDAR units have limitations: their cost is relatively high, and factors like heavy rain, fog, or very dark, absorbent surfaces can reduce detection effectiveness. Despite these challenges, ongoing improvements and reduced sensor costs lead to broader adoption. Overall, LiDAR in cars is crucial for enabling higher levels of driving automation by providing a reliable 3D sense of the road environment that underpins navigation and safety decisions \cite{b1}. 

\subsection{General issues with LiDAR data}\label{2b}

Despite its widespread utility, LiDAR data comes with several challenges that impact its efficiency and accuracy across applications. One significant issue is data quality degradation under adverse environmental conditions. Atmospheric elements such as rain, snow, fog, or dust can scatter or absorb laser pulses, leading to incomplete or noisy point clouds. For instance, dense fog significantly reduces signal return strength, impairing the accuracy of distance estimation and object detection, especially for automotive and aerial applications \cite{b1}. Additionally, surface reflectivity affects the accuracy of LiDAR returns. Dark or absorbent materials can absorb laser light, resulting in weak or missing returns. Highly reflective surfaces, on the other hand, may introduce multiple reflection points that misrepresent actual object geometry. Another issue is occlusion and line-of-sight limitations. Since LiDAR sensors rely on direct reflection from surfaces, objects obstructed by others will not appear in the point cloud, which can compromise situational awareness in navigation systems. Moreover, in urban or forested environments, multiple returns from complex geometries can complicate data interpretation, especially when isolating ground points from vegetation or built structures \cite{b12}. From a computational standpoint, large data volume is a bottleneck. High-resolution LiDAR surveys can generate billions of 3D points, resulting in datasets that require extensive storage and processing capabilities. Efficient filtering, segmentation, and classification algorithms are necessary to extract meaningful information, but these tasks remain computationally intensive and often require domain-specific tuning. These combined limitations highlight that while LiDAR is a powerful tool, its data requires careful handling, correction, and contextual interpretation to ensure reliable downstream applications.

\subsection{Security issues with LiDAR data}\label{2c}

LiDAR sensors in autonomous vehicles are vulnerable to adversaries manipulating the 3D point cloud data. Attackers can inject fake objects \cite{b13, b14} or erase real ones, creating phantom or missing obstacles that mislead the vehicle. One recent attack used only 20 added points to conjure a phantom car, fooling a state-of-the-art detector with 89\% accuracy \cite{b14}. Conversely, a vanishing attack made an autonomous vehicle miss a car in its path \cite{b15}. These attacks exploit weaknesses in deep learning-based 3D object detectors. Voxel-based detectors are especially vulnerable to point injection than point-based models \cite{b13}.
Meanwhile, point-based and graph neural network (GNN) detectors can be deceived by small, carefully crafted point perturbations, leading to detection errors. Attackers could also tamper with LiDAR data in transit or replay old point clouds to compromise data integrity \cite{b16}. A study \cite{b17} highlighted that adversarial point cloud attacks degrade model accuracy by over 40\% through strategic point perturbations. These vulnerabilities stem from inherent limitations in DNNs’ robustness to input distortions and the lack of adversarial training frameworks tailored for 3D data. To mitigate these risks, defenses are being developed at multiple levels. Sensor manufacturers have added hardware safeguards like randomizing laser pulse timing and embedding unique pulses to foil predictable spoofing \cite{b15}. Cryptographic watermarks can be embedded in point clouds at the data level to detect tampering \cite{b16}. \autoref{tab:my_label} lists relevant literature that addresses security issues in LiDAR datasets.
\\
Additionally, adversarial training using perturbed LiDAR datasets and real-time anomaly detection algorithms may mitigate spoofing risks. However, the evolving sophistication of attacks demands continuous updates to defense protocols and sensor hardware. Addressing these challenges is critical to ensuring the reliability of AVs, as unresolved vulnerabilities could lead to catastrophic safety failures. Future research must focus on standardized benchmarks for evaluating 3D model robustness and developing privacy-preserving LiDAR data processing techniques to balance security with performance.

\begin{table}[h]
    \centering
    \caption{A table highlighting relevant work that addresses security issues in LiDAR}
    \begin{tabularx}\linewidth{|p{1.8cm}|X|}
    \hline
    Study reference number     &  Description of the study\\
    \hline
    \cite{b13}     &      Demonstrates the generation of adversarial objects that are not detected by real-world LiDAR systems.\\
    \hline
    \cite{b14}    &       Demonstrates a highly deceptive adversarial obstacle generation algorithm against deep 3D detection models to mimic fake obstacles within the effective detection range of LiDAR.\\
    \hline
    \cite{b18}    &       Describes the vulnerability of current LiDAR-based perception architectures and discusses ignored occlusion patterns in LiDAR point clouds that make self-driving cars vulnerable to spoofing attacks.\\
    \hline
    \cite{b19}    &       A review of the various physical adversarial and spoofing attacks that LiDAR sensors are susceptible to. \\
    \hline
    \cite{b20}    &       A data hiding-based method is proposed to insert a binary watermark into the LiDAR data at the sensing layer, which is used for integrity verification and tamper detection and localization at the decision-making unit to prevent attacks.\\
    \hline
    \cite{b21}    &       Describes an attack method with the adversarial 3D object as the attack vector for Multi-Sensor Fusion in Autonomous Driving Vehicles.\\
    \hline
    \end{tabularx}
    \label{tab:my_label}
\end{table}

\subsection{Synthetic LiDAR data}\label{2d}
There are several issues when it comes to data generation and usage. While 2D image data is plentiful, LiDAR point cloud data is difficult to generate, especially for hyperlocal regions. To get point cloud data for a particular region, it is necessary to plan a drive-through or a flyby through that region (based on requirements for a ground or aerial dataset). This requires special equipment, time, and significant expenditure. While specific examples of datasets exist, such as the one described in \cite{b22}, the subject region is a university campus. Additional parameters such as traffic, safety, pedestrians, weather, and time of day also influence driving conditions and thus must be considered. Due to time, cost, and project constraints, dataset owners may often choose not to share their data with other users and researchers.
\\
Additionally, point cloud data is sensitive from a military and strategic perspective \cite{b23}; hence, sharing such datasets over the internet, where anyone can access them, is not always advisable. With the advent of deep learning and AI-backed applications such as autonomous driving \cite{b24} and image and data segmentation \cite{b25}, a large amount of data is required to train these models. Small, single-run datasets are not enough for training models. \autoref{tab2} lists recent and relevant research that describes frameworks for creating synthetic datasets. Various approaches, such as using specialized simulators, computer games, aerial photography, and GIS information sources, have been used by researchers to create synthetic datasets.

\begin{table}[h]
    \centering
    \caption{Relevant research that describes frameworks for creating synthetic datasets.}
    \begin{tabularx}\linewidth{|p{1.8cm}|X|}
    \hline
    Study reference number  &   Description of synthetic data generation method\\
    \hline
    \cite{b26}    &   Proposes a method for reconstructing scene geometry using LiDAR information to generate realistic synthetic point clouds as if they were captured by real devices.  \\
    \hline
    \cite{b27}     &  Using the CARLA simulator to extract LiDAR point clouds with ground truth annotations automatically{\cite{b28}}. \\
    \hline
    \cite{b29}    &   A synthetic 3D aerial photogrammetry point cloud dataset was created using GIS sources as a base layout.\\
    \hline
    \cite{b30}    &   Framework to rapidly create point clouds with accurate point-level labels from a computer game.\\
    \hline
    \cite{b31}    &   A modified CARLA simulator tool was created to generate synthetic semantic LiDAR scans.\\
    \hline
    \cite{b32}    &   A precise synthetic image and LiDAR dataset for autonomous vehicle perception was created using a custom simulator and GTA V computer game.\\
    \hline
    \cite{b33}    &   An automated synthetic LiDAR dataset generation workflow using the SVL simulator {\cite{b34}}.\\
    \hline
    This study  &   Uses a simulated environment in CoppeliaSim with several customizable objects and vehicle-mounted sensors to collect LiDAR and vision data.\\
    \hline 
    \end{tabularx}
    \label{tab2}
\end{table}

\section{Dataset Generation Pipeline}\label{3}
\begin{figure}[h]
    \centering
    \includegraphics[width=0.9\linewidth]{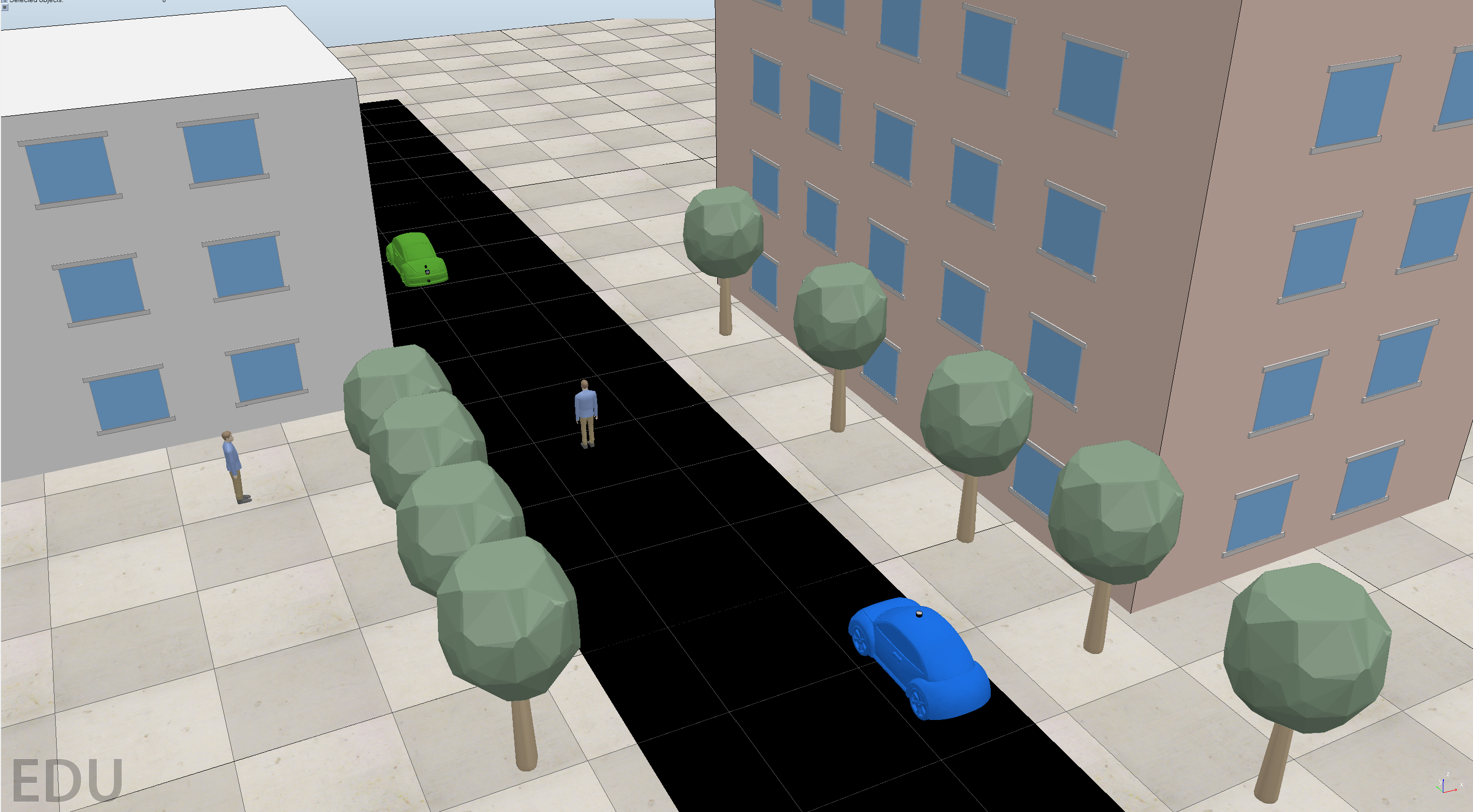}
    \caption{A view of the created simulation environment featuring trees, pedestrians, and vehicles as the primary objects.}
    \label{fig2}
\end{figure}

The data generation pipeline is created using the CoppeliaSim platform \cite{b35} and its Python API to collect, visualize, and process the data. The simulator is widely used in education, research, and industry applications. It primarily uses Python and Lua scripts and supports C++ and MATLAB through extensions and APIs. Several physics simulation libraries, such as Bullet \cite{b36} Newton Game Dynamics, are used to perform rigid body simulations. Meshes can be used to build objects and scenes. Instances of the environment were reused from previous experiments in the simulation environment conducted by the authors \cite{b37}. The prior work demonstrated the benefits of conducting UAV experiments in high-fidelity, dynamic environments. A combination of externally imported mesh-based objects and default available models was used in the simulation environment. An urban scenario was created with buildings, trees, cars, and people. The application is an autonomous driving vehicle scenario. \autoref{fig2} shows views of the created simulation environment and its different objects.

\begin{figure}[h]
    \centering
    \includegraphics[width=0.9\linewidth]{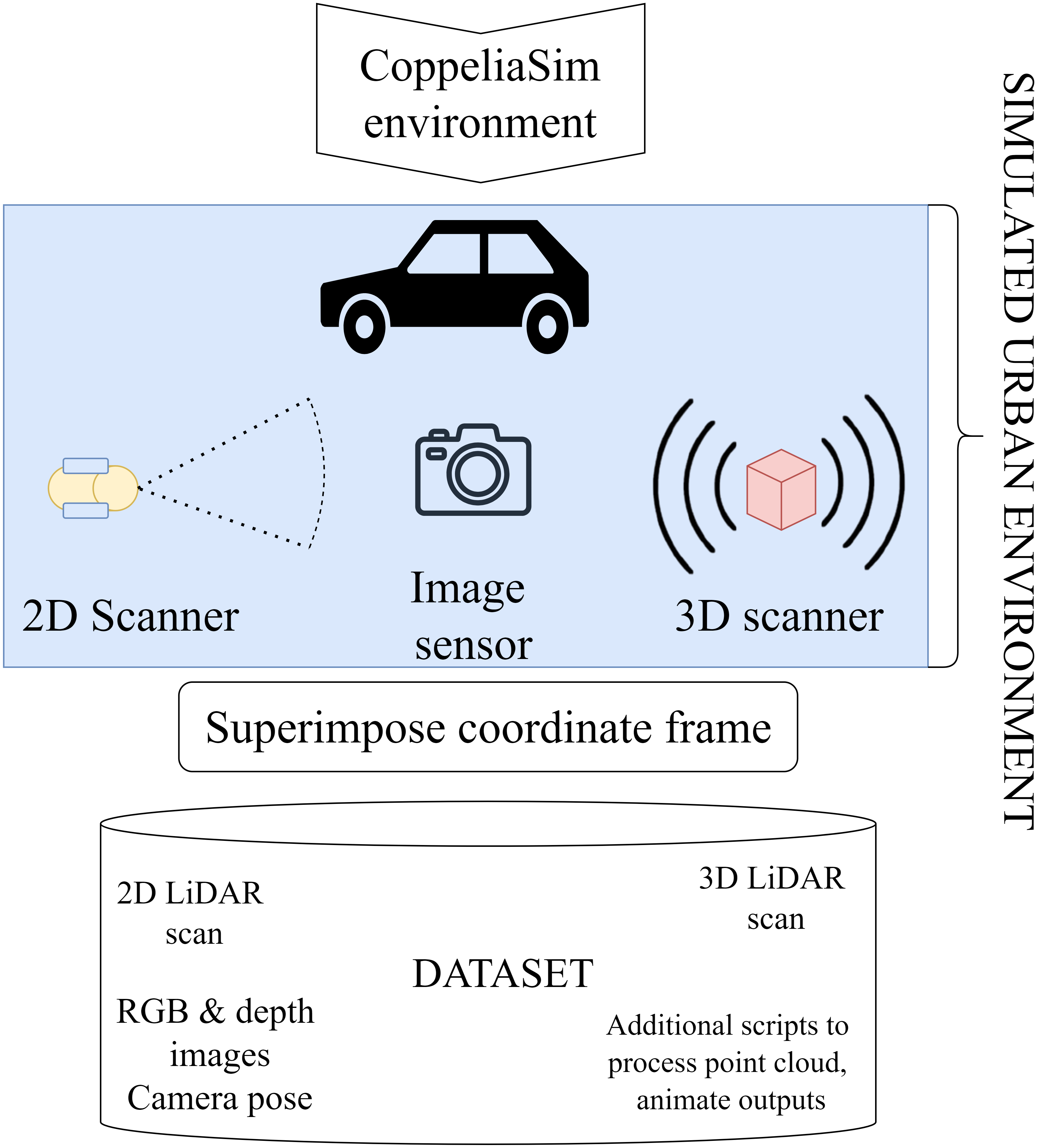}
    \caption{Workflow of the simulation environment.}
    \label{fig3}
\end{figure}

\begin{figure}[h]
    \centering
    \includegraphics[width=0.6\linewidth]{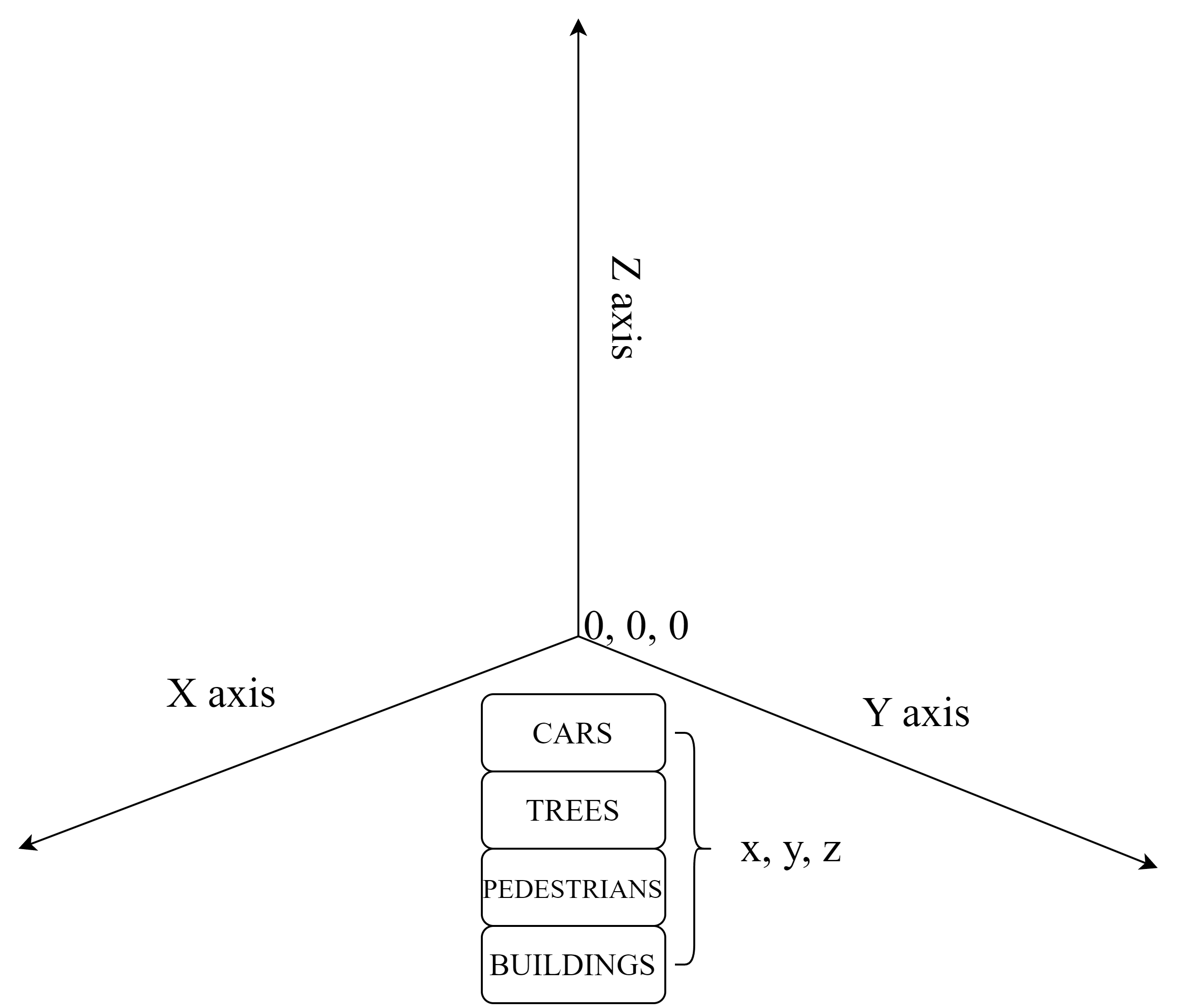}
    \caption{Three-axis coordinate frame and the presence of objects within the simulation environment.}
    \label{fig4}
\end{figure}

\autoref{fig3} shows the workflow used to create the synthetic data generation pipeline. The vehicle environment and the sensors are created and defined within the CoppeliaSim simulation environment. A local coordinate frame is created to reference the collected data. Each sensor is mounted in unique positions within the vehicle. The 2D scanner and the image sensor are mounted on the vehicle's front fender, and the 3D scanner is mounted on the rooftop. Each sensor collects and stores its data in associated folders in multiple formats. Additional scripts are present to combine, process, visualize, and animate the data as desired. The simulation uses a local coordinate frame to assign coordinates to the objects in the environment and perform calculations for sensor data collection, fusion, and processing. \autoref{fig4} shows a 3-axis view of the simulated world. Each object within the environment is assigned coordinates according to its placement and dimensions. This enables the sensor data that is collected to have location labels. 

\begin{figure}[h]
    \centering
    \includegraphics[width=0.8\linewidth]{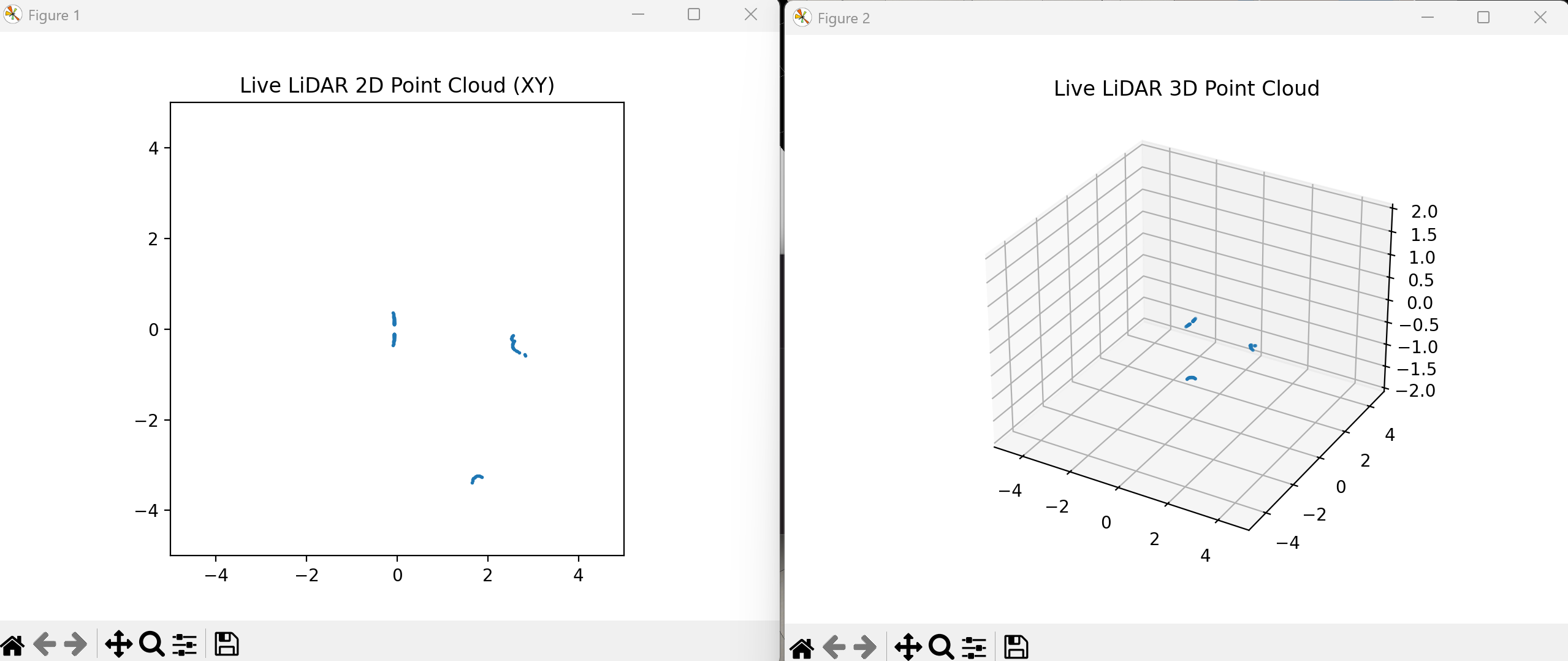}
    \caption{Live viewports for the 2D and 3D LiDAR outputs.}
    \label{fig5}
\end{figure}

Using the Python API, an instantaneous view of the LiDAR point cloud is also enabled. \autoref{fig5} shows one viewport taken when a simulation instance was running. 

\section{Dataset Description and Results}
A comprehensive sensor dataset is produced directly from pipeline creation and execution in the simulation environment. \autoref{tab3} and \autoref{tab4} describe the use cases of the generated data and the data fields in detail.

\begin{table}[ht]
    \centering
    \caption{Table highlighting generated data and use case scenarios.}
    \begin{tabularx}\linewidth{|p{1.4cm}|p{1.4cm}|p{1.4cm}|X|}
    \hline
    Type of data    &	Description   &   Final data collected &  Use case\\
    \hline
    2D LiDAR point cloud    &   Experimental sensor fitting &   2D planar scan  &   2D SLAM, scene mapping, fast persistent object detection\\
    \hline
    3D LiDAR point cloud    &   Primary data collection sensor  &   Full perspective 3D point cloud &   3D mapping, localization, obstacle detection\\
    \hline
    RGB images  &   Primary data collection sensors     &   Color images    &   Visual feature extraction, image segmentation, sensor fusion\\
    \hline
    Depth images    &   Primary data collection senor   &   Grayscale images    &   Dept aware perception, scene understanding, 3D reconstruction\\
    \hline
    Camera pose     &   Secondary data derivation based on primary sensors      &   Global position and orientation at each frame   &   Ground truth trajectory evaluation, localization, odometry, multiple frame alignment\\
    \hline
    \end{tabularx}
    \label{tab3}
\end{table}

\begin{table}[h]
    \centering
    \caption{Description of CSV fields of the point cloud data file.}
    \begin{tabularx}\linewidth{|p{1cm}|p{1cm}|p{0.7cm}|X|p{1.3cm}|}
    \hline
    Column name     &      Type     &   Units   &   Description     &   Significance\\
    \hline
    timestamp   &	ISO 8601 string   &   N/A     &   The exact time when the frame was captured      &   Allows temporal ordering, data synchronization\\
    \hline
    Frame	& Integer	&   N/A     &   Sequential frame index (starts from 0 or 1) &	Connects all data modalities per frame\\
    \hline
    x	&     Float   &   meters  &   X position of vehicle/camera in world  &    coordinates	Spatial location for trajectory\\
    \hline
    Y	&     Float   &   meters  &   Y position of vehicle/camera in world coordinates   &   {----}\\
    \hline
    Z	& Float   &   meters  &   Z position of vehicle/camera in world coordinates   &   {----}\\
    \hline 
    alpha   &     Float     &   radians     &   Rotation about X-axis (roll)    &   Orientation needed for 6-DoF pose\\
    \hline
    beta	&     Float	   &   radians     &   Rotation about Y-axis (pitch)   &   Orientation needed for 6-DoF pose\\
    \hline
    gamma	&     Float   &   radians     &   Rotation about Z-axis (yaw)     &   Orientation needed for 6-DoF pose\\
    \hline

    \end{tabularx}
    \label{tab4}
\end{table}

\begin{figure*}[t]
    \centering
    \includegraphics[width=0.9\linewidth]{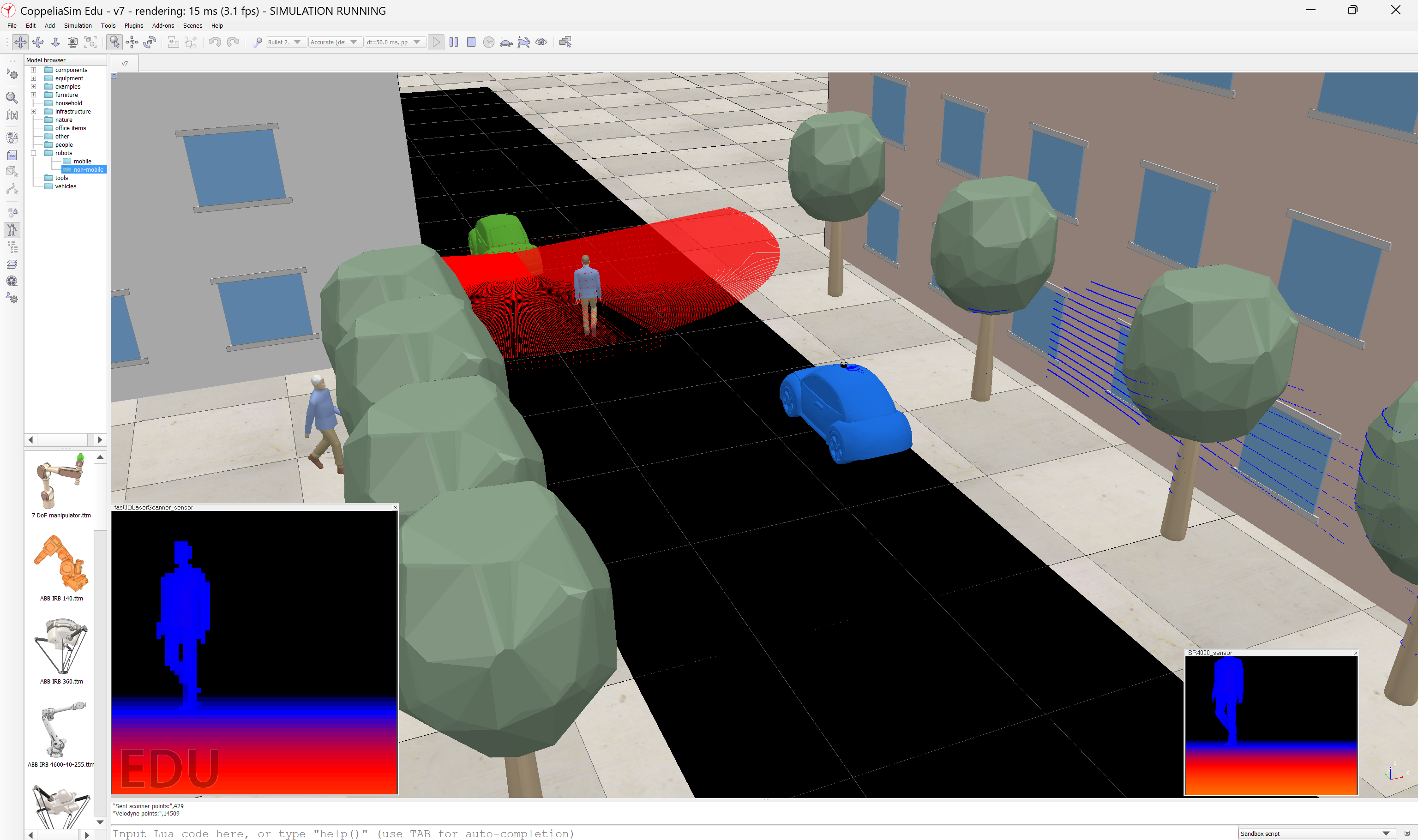}
    \caption{A screenshot of the simulation pipeline in progress showing detection viewports and LiDAR scanner projections.}
    \label{fig6}
\end{figure*}

\begin{figure}[h]
    \centering
    \includegraphics[width=0.9\linewidth]{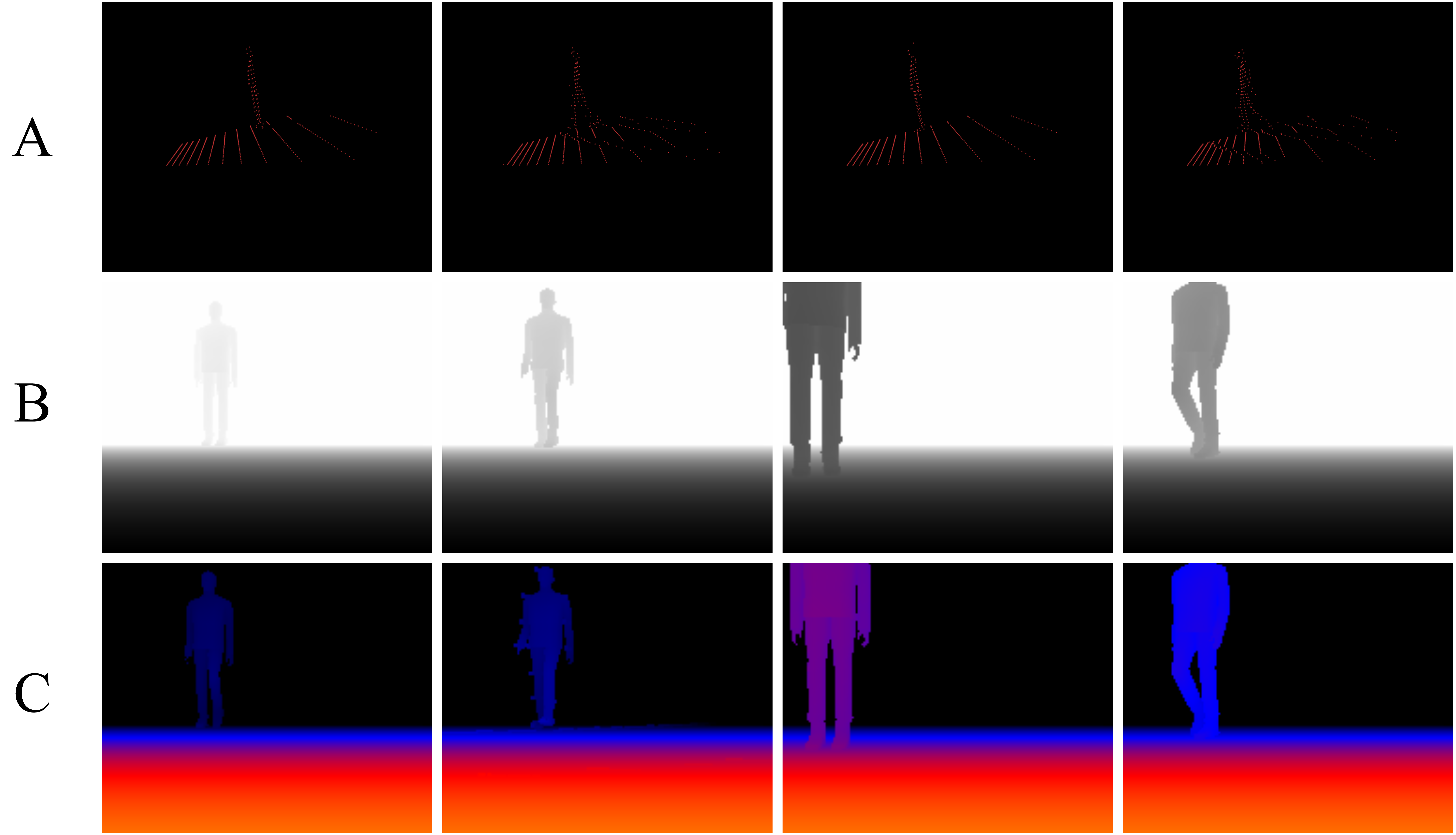}
    \caption{Different viewports showing A) Animated point cloud, B) Output from the depth sensor, and C) Output from the RGB sensors.}
    \label{fig7}
\end{figure}

\autoref{fig6} shows a screenshot of the simulation in progress. Multiple LiDAR projections can be seen, and their detections are recorded via view port windows. \autoref{fig7} shows outputs from the multiple viewport windows that record sensor outputs.

\begin{figure}[h]
    \centering
    \includegraphics[width=0.9\linewidth]{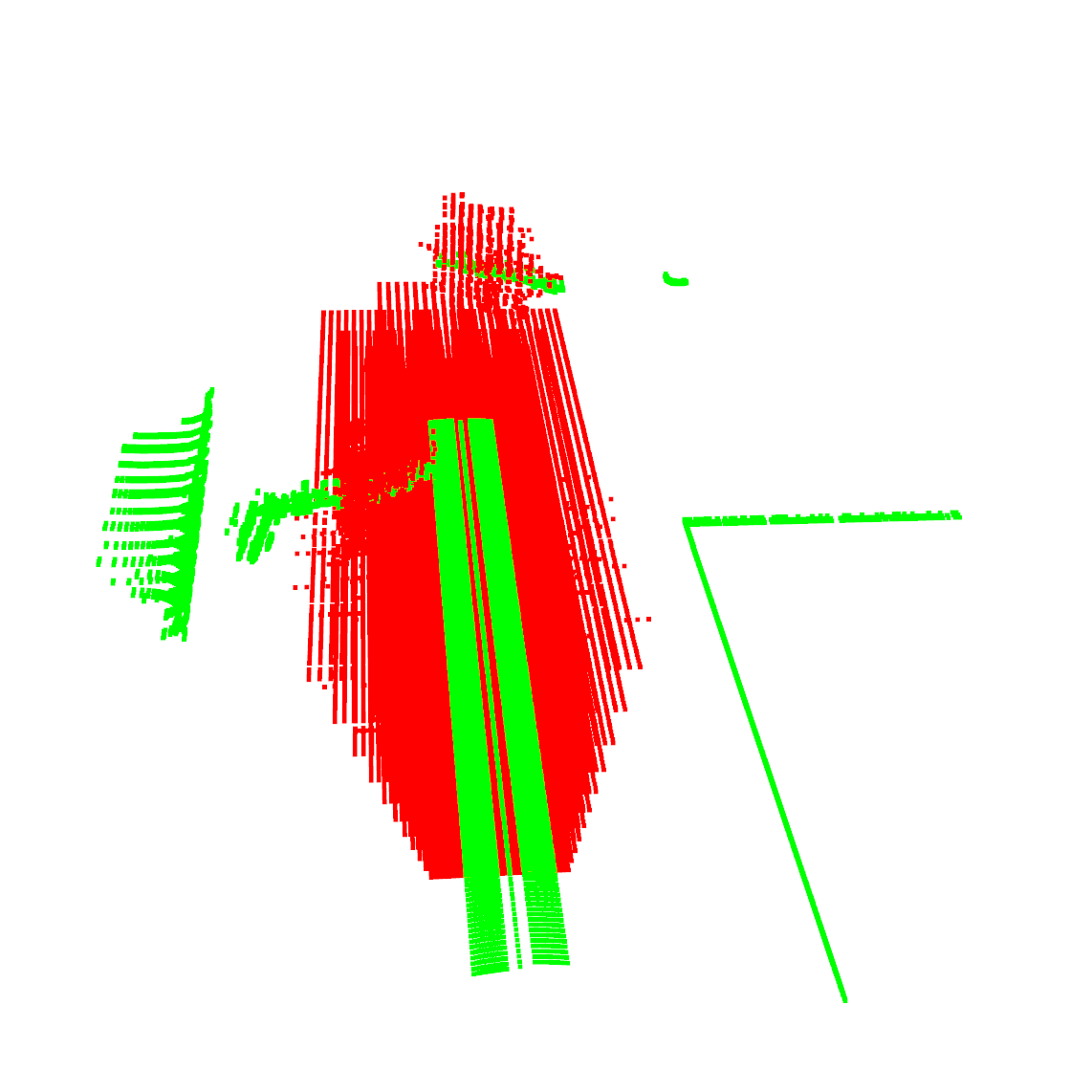}
    \caption{Aggregated point cloud from an oblique viewport for the 3D scanner.}
    \label{fig8}
\end{figure}

Point cloud data is stored in both pcd and ply formats. Additional scripts for aggregation and feature matching are used. The aggregation script ingests per-frame point clouds and the vehicle’s ground-truth pose to build a unified map. For each frame, it: (1) reads the corresponding .ply file; (2) looks up the vehicle’s 6-DoF pose (x, y, z, roll, pitch, yaw) from poses.csv; (3) applies the rigid-body transform to bring each point into the world coordinate frame; and (4) tags points with a modality-specific color. All transformed points are concatenated into one large Open3D PointCloud object, optionally down-sampled via a voxel grid, and then exported to both .ply and .pcd formats for compatibility with downstream tools. Finally, the combined cloud is rendered in an interactive Open3D window for visual inspection. However, several limitations apply. Memory usage can grow prohibitively large as concatenating thousands of high-density frames without on-disk streaming may exhaust system RAM. Second, the approach assumes perfect pose accuracy. Any calibration errors or drift in the simulated pose will introduce misalignments and ‘‘ghost’’ artifacts in the merged cloud. No outlier removal or noise modeling is performed, so spurious points (e.g., sensor dropouts or reflections) are carried through.
\\
Additionally, temporal information is lost in the final merged map, preventing dynamic-object filtering or sequence-based analysis. Finally, voxel down-sampling trades spatial fidelity for performance, which can obscure fine details. Addressing these issues would require chunked loading, pose refinement (e.g., ICP registration), statistical filtering, and preserving frame metadata for advanced filtering or temporal reconstruction. The aggregation is done because doing so provides a convenient way to examine the target map in varying levels of resolution. Sectional aggregations can be created by dividing the collected data and providing higher-density views of a particular object or region. \autoref{fig8} shows the aggregation from an oblique viewport.

The multi-sensor data enables testing of cross-modal sensor fusion techniques. Timestamped data can simulate real-time streaming and synchronization. Ground truth pose allows for SLAM validation and mapping benchmarks.

\subsection{Using advanced and variable scanners to increase dataset variability and fidelity}

\begin{figure}[h]
    \centering
    \includegraphics[width=0.9\linewidth]{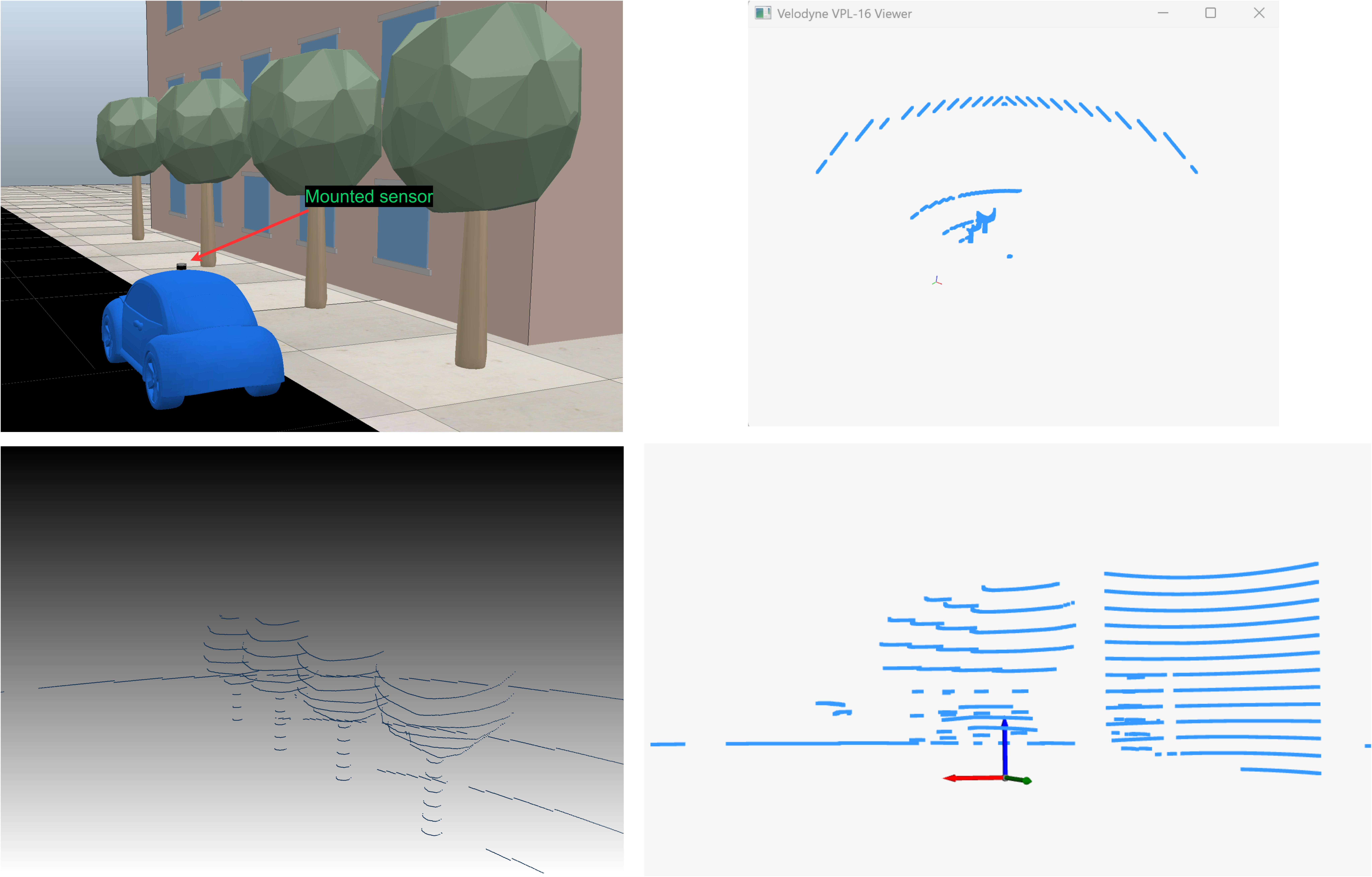}
    \caption{Different views of the simulation environment and produced dataset.}
    \label{fig9}
\end{figure}

\begin{figure}[h]
    \centering
    \includegraphics[width=0.9\linewidth]{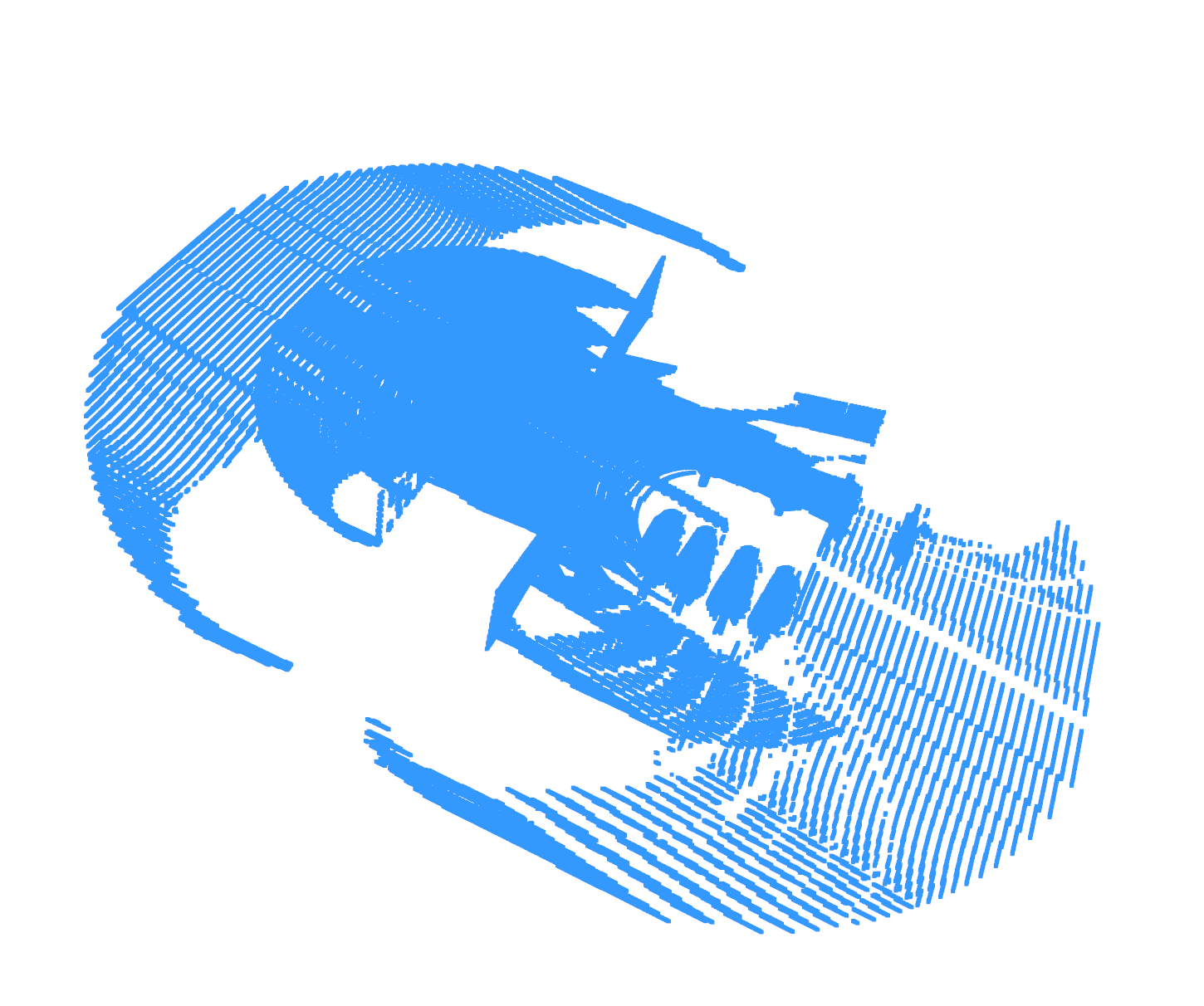}
    \caption{Aggregated point cloud for the complete simulation environment collected by the Velodyne sensor.}
    \label{fig10}
\end{figure}

A different car in the simulation was fitted with the Velodyne VLP-16 3D LiDAR sensor. This small puck-shaped sensor is cost-effective and has applications in the autonomous vehicles and robotics industry \cite{b38}. While it provides less accuracy and range than the Velodyne HDL series, the VLP has dual return capability, low power consumption, and a broad 360-degree horizontal and 30-degree vertical field of view. The maximum range is set at 100 meters. \autoref{fig9} shows the car with a mounted sensor (top left), a view of the live feed from the Velodyne VPL 16 (top right), and the resultant processed point clouds from two different angles and evaluated using third-party tools for coordinate frame orientation and imposition of GNSS coordinates if required (bottom left and right). \autoref{fig10} shows the complete aggregated point cloud for the Velodyne data. This was generated using the same method used for the aggregated point cloud from the primary LiDAR sensor, the method for which is described in the previous section.

\subsection{Feature matching for advanced tracking and visualization}
\begin{figure*}[t]
    \centering
    \includegraphics[width=0.9\linewidth]{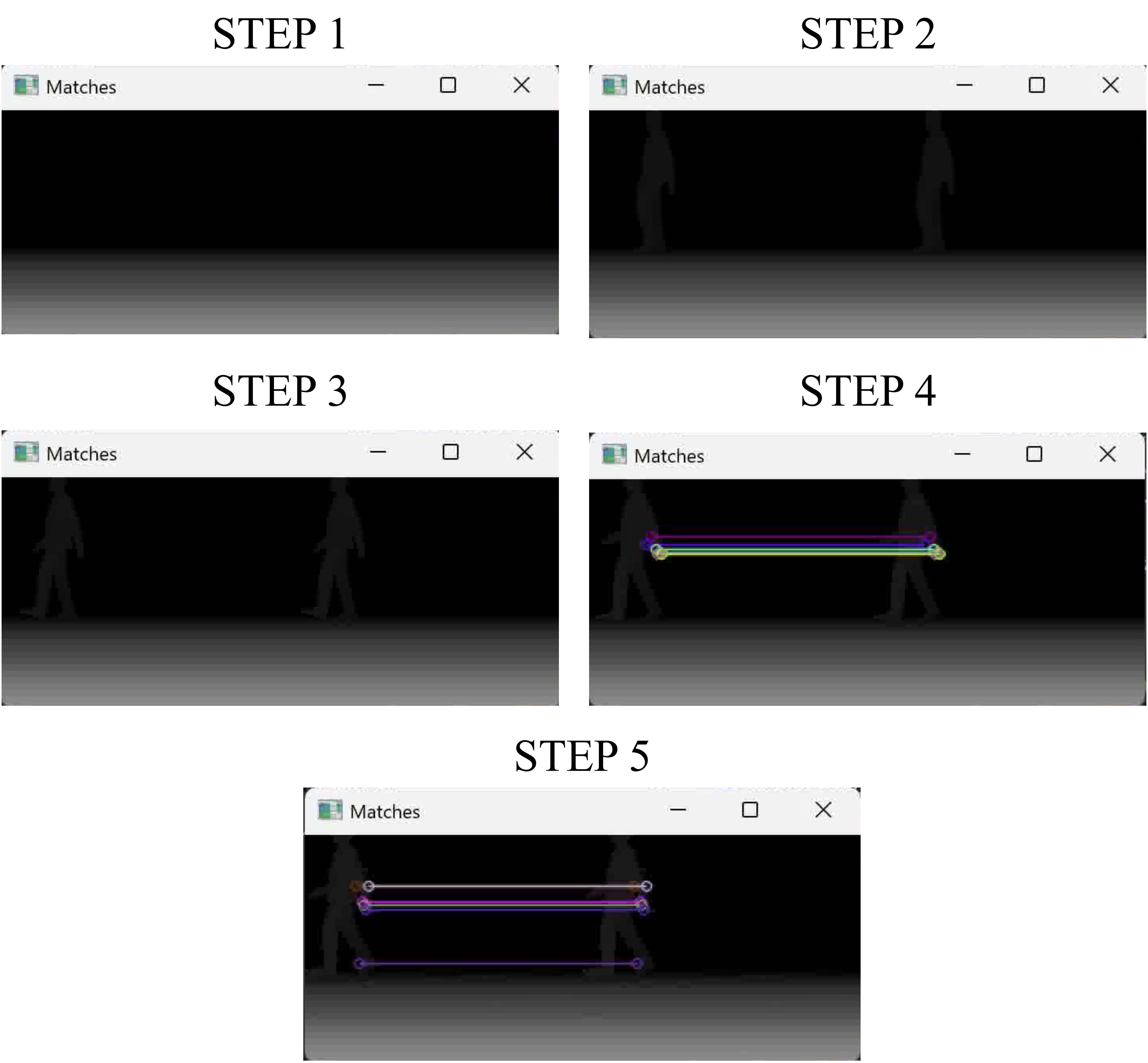}
    \caption{Feature matching and trajectory visualization using captured point cloud data.}
    \label{fig11}
\end{figure*}

Feature tracking and visual odometry (VO) are foundational pipelines for estimating camera motion from monocular image sequences. In this implementation, each RGB frame is converted to grayscale and processed with the ORB (Oriented FAST and Rotated BRIEF) detector to extract up to 500 robust key points and corresponding binary descriptors. Consecutive frames are matched using a brute-force Hamming-distance matcher with cross-check, yielding pairs of pixel correspondences. Although the demonstration script visualizes matches, a full VO system would compute the essential matrix via RANSAC (Random Sampling And Consensus) to reject outliers and decompose it into rotation and translation components (up to scale) using known camera intrinsics. Estimated relative poses can then be concatenated to form an incremental trajectory. \autoref{fig11} shows a brief output from the generated data that extracts features from a detected person.

This approach has several limitations. First, monocular VO suffers from scale ambiguity; without external information (e.g., stereo disparity, LiDAR depth, or GPS), absolute translation magnitude cannot be recovered. Second, feature-based methods are sensitive to lighting changes, motion blur, and low-texture regions where insufficient key points are detected. Third, the drift accumulates over time since errors in each pose estimate compound without global loop closure or map optimization. Finally, the simplified script used here omits robust outlier rejection beyond RANSAC and does not account for rolling shutter distortions or lens radial distortion. While ORB-based VO is lightweight and real-time capable, it requires careful calibration, complementary sensors for scale, and advanced SLAM techniques to produce globally consistent trajectories.

\section{Future Work}
This work has several limitations as well as directions for further research. Although the simulator supports the integration of weather effects \cite{b39}, it is currently not integrated into this workflow. LiDAR performance deteriorates in conditions such as rain, fog, and snow. While random or logic-based simulated scattering effects can be applied, weather effects are required for better realism effects and the best replication of the synthetic data to match real-world datasets. Additionally, real-world sensors suffer from range noise, multi-path detections, missed dropouts, and mixed pixel returns. Algorithms trained/tested on clean simulated data can overfit and fail when encountering real sensor imperfection. Synthetic noise models like Gaussian range noise and random dropouts should be added when designing the sensors in the simulation to represent errors better.
\\
Additionally, advancements in simulator tools themselves can lead to improvements in the manner in which simulations are performed. By default, the sensor in the pipeline only generates x, y, and z positions. Currently, the only way to simulate intensity values in CoppelaSim for LiDAR returns is by modifying the material properties to return specific values or generate pseudo-intensity maps. Uniform textures on object models such as buildings and roads can produce simple and uniform reflectivity. A possible solution is to use higher realism in 3D scene textures and surface materials. However, this increases the time required to design these resources during the development stage of the target environment. The simulator should also be capable of handling multiple high-resolution models. 

\subsection{Using real-world locations in simulators to create synthetic datasets}
\begin{figure}[h]
    \centering
    \includegraphics[width=0.9\linewidth]{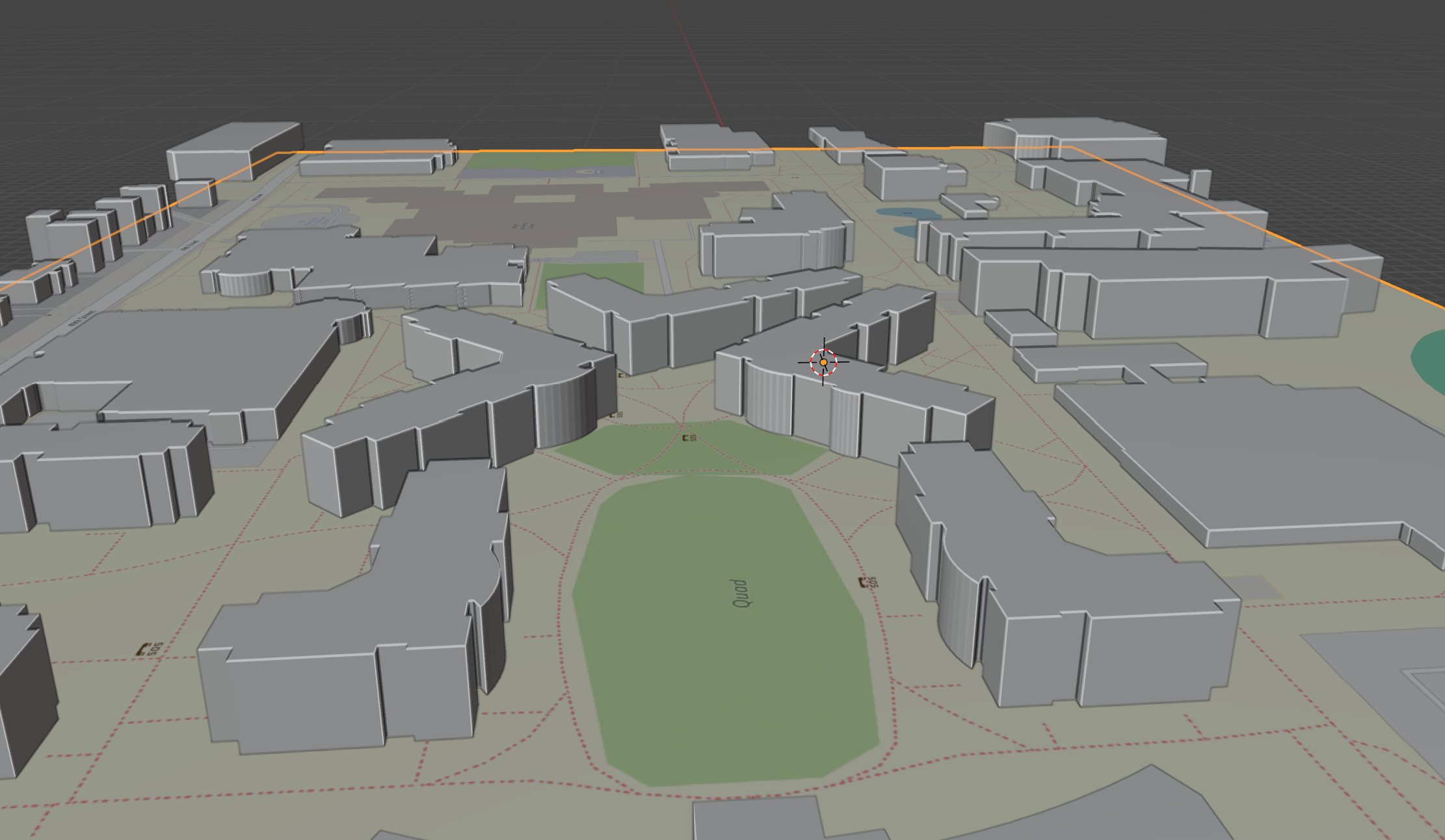}
    \caption{A mesh model of the Old Dominion University campus (Norfolk, VA) will be used in the simulation model.}
    \label{fig12}
\end{figure}

Simulation environments are often limited to a designer’s capacity to imagine and create test environments for deploying robots and generating synthetic datasets. Most current simulation environments use simplified, repetitive obstacle designs to test crewed and uncrewed vehicle deployments \cite{b37}. \autoref{fig12} shows a mesh model of the Old Dominion University (Norfolk, Virginia) campus derived using OSM data. This 3D reconstruction of the campus imported in the simulator provided a reliable environment to deploy autonomous vehicles and create synthetic datasets using the pipeline described above. Since real-world locations have urban planning and code regulations enforced, this adds a facet of dimensionality to the synthetic data.

For example, such models imposed to scale and subjected to the data generation pipeline for LiDAR and image data collection can provide high-resolution data for applications such as using automated robots on campus premises for various functions such as food delivery security \cite{b40} and automated tour guides \cite{b41}.

\subsection{Using the dataset to build optimized techniques to detect security vulnerabilities in LiDAR sensors}
\begin{figure}[h]
    \centering
    \includegraphics[width=0.9\linewidth]{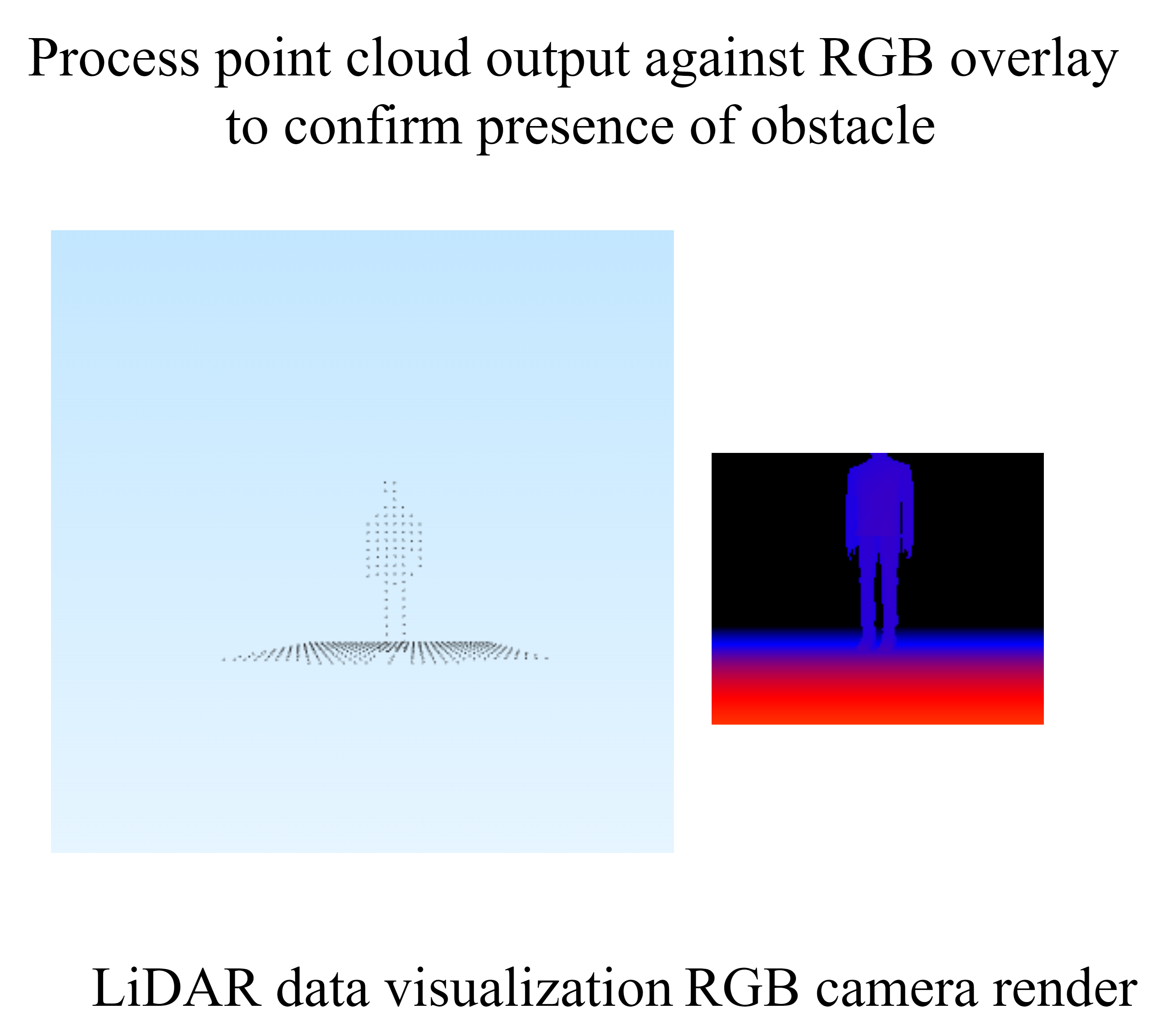}
    \caption{Future work to create an optimized confirmation pipeline to compare LiDAR and image sensor outputs as overlays to confirm the presence of obstacles and prevent spoofing and insertion attacks.}
    \label{fig13}
\end{figure}

Current work by the authors examines the overlapping of data through optimized overlapping techniques to fuse image data and point cloud data outputs to produce confirmation of obstacle presence in front of the sensors. \autoref{fig13} shows the 3D scanner and the image data side-by-side for the same timestamp. An overlay and point-matching technique to draw established edges to detect objects can help confirm the validity and presence of the object placed in front of the vehicle. This could effectively overcome some of the security concerns of LiDAR data outlined in previous sections.

\section{Conclusions}
The presented workflow establishes a reproducible method for generating high-fidelity synthetic LiDAR data within a unified simulation environment. It delivers multimodal datasets by combining ToF LiDAR, 2D scanning, and image sensors on a modular vehicle platform. These datasets can serve as robust testbeds for evaluating SLAM algorithms under controlled conditions, offering insights into spatial resolution trade-offs and synchronization accuracy. They can also support sensor security research by embedding security perturbation scenarios, such as adversarial point injection and spoofing attacks; the framework enables a systematic assessment of defense mechanisms against emerging threats. The simulation pipeline can produce large-scale point clouds in a range of environments. Including synchronized RGB and depth imagery further supports cross-modal fusion research, enhancing the development of perception systems for autonomous vehicles and UAVs. Limitations identified include the absence of integrated weather models, simplified reflectivity assumptions, and computational overhead associated with processing high-density point clouds. Addressing these gaps through advanced noise modeling, textured scene materials, and optimized data management will improve realism and scalability. Future extensions will incorporate real-world terrain reconstructions from geographic data sources and dynamic weather effects to emulate visibility degradation. The modular design of the workflow ensures that new sensor types and environmental conditions can be readily integrated. By releasing comprehensive documentation, example scenarios, and benchmark scripts alongside the codebase, we aim to accelerate the adoption of synthetic data for perception research and sensor security evaluation. This work provides a foothold for creating a versatile, extensible framework for synthetic LiDAR dataset generation, paving the way for more resilient perception systems, and advances the state-of-the-art in autonomous navigation and sensor protection.

\section{Acknowledgment}
This work was supported in part by the
Coastal Virginia Center for Cyber Innovation (COVA CCI) and the Commonwealth Cyber
Initiative (CCI), an investment in the advancement of cyber research \& development,
innovation, and workforce development.

\end{document}